\documentclass{article}

\usepackage{PRIMEarxiv}

\usepackage[utf8]{inputenc} 
\usepackage[T1]{fontenc}    
\usepackage{hyperref}       
\usepackage{url}            
\usepackage{booktabs}       
\usepackage{amsfonts}       
\usepackage{amsmath, amssymb}
\usepackage{nicefrac}       
\usepackage{microtype}      
\usepackage{lipsum}
\usepackage{fancyhdr}       
\usepackage{graphicx}       
\usepackage[section]{placeins}
\graphicspath{{media/}}     
\usepackage{cleveref}
\usepackage{multirow} 
\usepackage{pdfpages}

\title{StructCore: Structure-Aware Image-Level Scoring for Anomaly Detection
\thanks{\textit{\underline{Citation}}: 
\textbf{StructCore: Structure-Aware Image-Level Scoring for  Anomaly Detection}} 
}

\author{
  Joongwon Chae$^{1,2}$, Lihui Luo$^1$, Yang Liu$^1$, Runming Wang$^1$, Dongmei Yu$^3$, \\
  \textbf{Zeming Liang$^4$, Xi Yuan$^4$, Dayan Zhang$^4$, Zhenglin Chen$^4$, Peiwu Qin$^{4*}$, Ilmoon Chae$^{2*}$} \\
  \\
  $^1$Tsinghua University Shenzhen International Graduate School, Shenzhen, China \\
  $^2$Ratel Soft \\
  $^3$Affiliated Fifth Hospital, Wenzhou Medical University, Wenzhou, Zhejiang, China \\
  $^4$Chinese Medicine Guangdong Laboratory \\
  \texttt{cai-zy24@mails.tsinghua.edu.cn} \\ 
  \thanks{*Corresponding author}
}

\begin{document}
\maketitle

\begin{abstract}
Max pooling is the de facto standard for converting anomaly score maps into image-level decisions in memory-bank-based unsupervised anomaly detection (UAD).
However, because it relies on a single extreme response, it discards most information about how anomaly evidence is distributed and structured across the image, often causing normal and anomalous scores to overlap.
We propose \textbf{StructCore}, a training-free, structure-aware image-level scoring method that goes beyond max pooling.
Given an anomaly score map, StructCore computes a low-dimensional structural descriptor $\phi(S)$ that captures distributional and spatial characteristics, and refines image-level scoring via a diagonal Mahalanobis calibration estimated from train-good samples, without modifying pixel-level localization.
StructCore achieves image-level AUROC scores of 99.6\% on MVTec AD and 98.4\% on VisA, demonstrating robust image-level anomaly detection by exploiting structural signatures missed by max pooling.
\end{abstract}

\section{Introduction}

Unsupervised Anomaly Detection (UAD) is a core component of industrial visual inspection, where anomalies are rare, diverse, and costly to annotate. In many production settings, only normal samples are available for training, and the primary operational requirement is a reliable \emph{image-level} accept/reject decision under strict latency and memory budgets \cite{bergmann2019mvtec}. While pixel-level localization is important, the final image-level decision rule often determines practical system reliability.

Despite diverse modeling choices, many UAD approaches share a common interface: they produce a dense anomaly score map and then reduce it to a single image-level score for decision making. This pattern appears in reconstruction-based methods \cite{zavrtanik2021draem}, teacher--student discrepancy methods \cite{deng2022anomaly,wang2021stpm}, density/flow-based models \cite{rudolph2021same,gudovskiy2022cflow}, and memory-bank / nearest-neighbor pipelines that are widely adopted in industrial inspection due to their simplicity and strong performance \cite{defard2021padim,cohen2020spade,roth2022patchcore}. However, the final aggregation step is still dominated by a long-standing convention: \emph{max pooling} (or minor variants) over patch-wise anomaly scores \cite{roth2022patchcore,defard2021padim,cohen2020spade}. Max pooling is computationally cheap, but it relies on a single extreme response and discards most information about how anomaly evidence is distributed and structured across the image.

We argue that this reduction can be suboptimal in practice. For subtle or spatially distributed defects, normal and anomalous images may overlap substantially in max-pooled scores. Moreover, with strong pretrained transformer representations, localized spurious peaks can dominate the maximum even when the global anomaly pattern contains decisive evidence \cite{dosovitskiy2021vit,oquab2023dinov2}. In other words, the maximum response is often not a sufficient statistic for deciding whether an image is anomalous. Human inspectors rarely judge an image based on a single hottest pixel; instead, they rely on \emph{structure}---how anomaly responses spread, concentrate, and organize spatially.

This perspective is further motivated by a counterintuitive observation in recent ViT-based reconstruction-style methods such as Dinomaly \cite{guo2025dinomaly}. By leveraging global context in Vision Transformers, these methods can produce highly refined normal feature maps and mitigate the classical ``identity mapping'' issue, which might suggest that image-level detection should approach a performance ceiling. Yet, in practice, simple max pooling can still fail to provide the expected separation in certain cases. We interpret this as a \emph{structural gap}: even when anomaly maps are high-quality, decisive evidence may reside not in a single extreme response but in the collective distributional and spatial structure of the map, which max pooling inherently ignores.

Motivated by this, we propose \textbf{StructCore}, a training-free, structure-aware image-level scoring module that goes beyond max pooling. Given an anomaly score map $S$, StructCore computes a compact, low-dimensional structural descriptor $\phi(S)$ that summarizes distributional and spatial signatures of the map. Using only train-good samples, we estimate normal statistics of $\phi(S)$ and compute a diagonal Mahalanobis calibration score to refine image-level decisions \cite{lee2018mahalanobis}. Importantly, StructCore refines image-level scoring \emph{without modifying pixel-level localization}: patch scores and anomaly maps remain unchanged.

Finally, real deployments increasingly demand scalability beyond the classical single-category setting, including multi-category and continual scenarios where new product types appear over time. While our focus is structure-aware image scoring, we additionally consider an optional lightweight routed inference setup---following prior routing mechanisms---to demonstrate that StructCore remains modular and compatible in scalable deployment settings \cite{chae2026gcr}. This routing component is included as a practical system configuration rather than a core contribution.

Our contributions are three-fold: (1) we analyze max pooling as a critical bottleneck in image-level decision making that discards informative structural evidence in anomaly score maps; (2) we propose \textbf{StructCore}, a training-free structure-aware image scoring module based on low-dimensional map descriptors and train-good statistical calibration, compatible with existing UAD pipelines \cite{lee2018mahalanobis}; and (3) we provide extensive ablations over representations, memory, and structural descriptors, demonstrating stable image-level gains even with a minimal low-dimensional $\phi(\cdot)$.
\begin{figure*}[t]
    \centering
    \includegraphics[width=\textwidth]{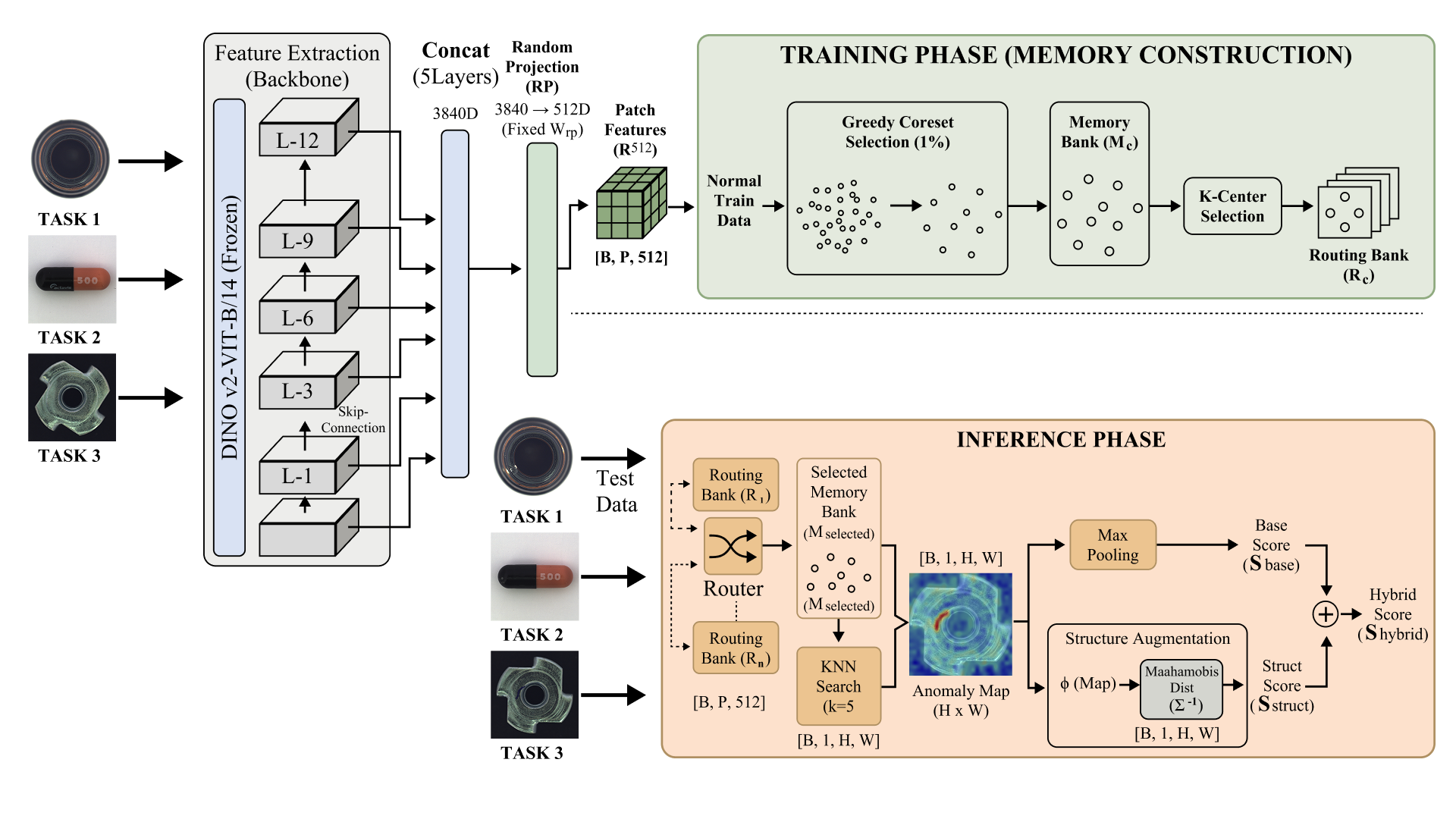}
    \caption{
    Overview of the StructCore framework.
    Normal images are encoded by a frozen DINOv2 ViT-B/14 backbone using multi-layer skip feature extraction.
    Patch features from multiple layers are concatenated and compressed via a fixed random projection, followed by greedy coreset selection to construct a category-specific memory bank.
    At inference time, an optional routing bank selects a relevant memory subset for efficient nearest-neighbor matching, producing an anomaly score map.
    While the base image-level score is obtained by conventional max pooling, StructCore augments it with a structure-aware score computed from a low-dimensional descriptor of the anomaly map, calibrated using statistics from train-good samples.
    StructCore refines image-level decisions without altering pixel-level localization.
    }
    \label{fig:overview}
\end{figure*}

\section{Related Work}
\label{sec:related}

Industrial visual anomaly detection (VAD) is commonly studied under a one-class protocol, where only normal images are available during training and the goal is to detect and localize defects at test time.
Benchmarks such as MVTec AD \cite{bergmann2019mvtec} and VisA \cite{zou2022spot} standardize evaluation by reporting both image-level detection and pixel-level localization.
Despite the diversity of model families, a large portion of practical industrial pipelines share a consistent interface: they first produce an anomaly score map and then reduce it to a single scalar image score for accept/reject decisions.
In this work, we target this interface explicitly.
Instead of modifying pixel-level anomaly maps (and thus localization behavior), we focus on improving the final \textbf{map-to-score} decision rule at image level.

Among training-free or training-light approaches, retrieval- and memory-bank-based methods are widely adopted due to their strong transferability from pretrained representations and low maintenance cost.
SPADE \cite{cohen2020spade}, PaDiM \cite{defard2021padim}, and PatchCore \cite{roth2022patchcore} compute anomaly maps via patch-wise matching in a pretrained feature space, often combined with memory-bank compression (e.g., coreset selection) and efficient nearest-neighbor search.
While these methods can yield high-quality anomaly maps, their final image-level decision is still typically dominated by max pooling (or minor variants such as top-$k$ pooling).
This choice is appealing for simplicity, but it can discard distributional and spatial evidence in the anomaly map, making image-level scores overlap between normal and anomalous samples even when localization is accurate.

The importance of aggregation from spatial maps to image-level decisions has been repeatedly emphasized in broader computer vision contexts. CAM-style analyses show that global pooling choices fundamentally affect what spatial evidence is preserved for recognition \cite{zhou2016cam}, and weakly-supervised localization and MIL-style formulations have explored alternatives to hard max pooling, including top-$k$ aggregation and class-wise pooling schemes \cite{durand2017wildcat}. Smooth approximations such as log-sum-exp pooling further illustrate that the aggregation operator is a design variable that controls the trade-off between peak sensitivity and evidence coverage \cite{yao2018multires}. These insights motivate a parallel question for anomaly detection: \emph{is the maximum response a sufficient statistic for deciding whether an image is anomalous?} This question becomes more pronounced with strong pretrained vision transformers \cite{dosovitskiy2021vit,oquab2023dinov2}, where normal images may contain localized spurious spikes while true anomalies can manifest as broader shifts or coherent structures across the map.

Beyond these pooling formulations, a broader line of work has shown that global aggregation can be interpreted as evidence summarization, and that retaining richer global statistics of spatial responses can improve robustness and generalization. In particular, second-order and covariance-based pooling methods demonstrate that incorporating global dispersion/heterogeneity statistics can preserve distributional structure that is largely discarded by first-order pooling alone \cite{lin2015bilinear,gao2019global,wang2020deep}. While these works are not designed for anomaly detection, they provide complementary motivation for our setting: when map-to-score reduction is unavoidable, leveraging compact global statistics can be preferable to relying on a single extreme response.

Several works in anomaly detection implicitly revisit the map-to-score issue through alternative aggregation or calibration strategies. Unified frameworks may employ top-$k$ or top-ratio matching to construct image-level anomaly scores \cite{lee2026uniformaly}, and recent test-time re-scoring methods aim to reduce false positives caused by noisy peaks by recalibrating image-level decisions without retraining \cite{li2024musc,he2025rareclip}. Compared to these approaches, StructCore formalizes a simple and fully training-free principle: summarize the anomaly map using a low-dimensional structural descriptor and calibrate it using train-good statistics via a diagonal Mahalanobis distance \cite{lee2018mahalanobis}. This directly compensates for the information loss of max pooling while keeping the anomaly map unchanged.

Our design is also connected to broader anomaly/OOD literature that motivates energy- and tail-aware scoring. Energy-based perspectives treat anomaly detection as thresholding an energy (or free-energy) score, and have been studied in classical anomaly detection settings \cite{zhai2016dsebm} as well as modern OOD detection, where energy scores provide a stable alternative to softmax confidence \cite{liu2020energy_ood}. Tail-based reasoning is similarly natural, as anomalies often correspond to rare events located in distribution tails. COPOD \cite{li2020copod} and ECOD \cite{li2022ecod} estimate tail probabilities for outlier scoring, while EVT-based open-set recognition methods such as OpenMax explicitly model extreme values for unknown detection \cite{bendale2016openmax}. A recent survey further systematizes these directions under generalized OOD detection, highlighting the recurring role of tail/extreme behavior as a discriminative cue across settings \cite{yang2024generalized}. These studies support the general usefulness of tail/extreme cues, but they mostly operate on scalar scores or latent distributions. In contrast, StructCore applies tail-sensitive statistics directly to anomaly \emph{maps} and uses them for image-level calibration.

Finally, capturing spatial organization beyond peak magnitude relates to classical image-processing measures such as total variation (TV), which quantifies spatial roughness and discontinuity and has been widely used since the ROF model \cite{rudin1992rof}. Practical implementations and analyses of ROF-style TV regularization further clarify its interpretation as a map-level roughness measure and provide efficient algorithms \cite{getreuer2012rudin}. Such measures are particularly suitable for anomaly maps, where a key ambiguity is whether high responses form a coherent defect region or scattered noise. Moreover, region-shape descriptors based on perimeter--area relations provide complementary cues of compactness versus fragmentation for thresholded high-score regions, and have been studied as compactness/circularity measures in pattern analysis \cite{montero2009state}. By combining dispersion, tail mass, and spatial structure into a compact descriptor, StructCore provides a lightweight yet effective image-level scoring module that is compatible with memory-bank detectors and orthogonal to system-level scalability mechanisms such as routing or mixture-of-experts designs \cite{meng2024moead,gu2025anomalymoe,chae2026gcr}.

\section{Method}
\label{sec:method}

\subsection{Overview}
\label{sec:method:overview}
Our approach augments a PatchCore-style memory-based UAD pipeline~\cite{roth2022patchcore} with StructCore, a module that improves image-level scoring without additional gradient-based training.
During fitting, for each category $c\in\mathcal{C}$ we build (i) a compact scoring memory bank $\mathbf{M}_c$ via coreset selection, (ii) a routing prototype bank $\mathbf{R}_c$ for multi-category binding, and (iii) normal statistics required by StructCore.
At inference, we route an input to a single category $\hat{c}$ using $\mathbf{R}_c$, query only $\mathbf{M}_{\hat{c}}$ to produce an anomaly map, and then apply StructCore to quantify structural deviation of the map and refine the final image-level score.
StructCore changes only the image-level score and does not modify the anomaly map, preserving pixel-level localization behavior.

\subsection{Skip-layer Multi-layer ViT Features}
\label{sec:method:features}
We use patch tokens from a frozen DINOv2 ViT-B/14 backbone~\cite{oquab2023dinov2,dosovitskiy2021vit}.
Since memory-based UAD relies on patch-wise retrieval, we discard the CLS token~\cite{roth2022patchcore,defard2021padim,cohen2020spade}.
Instead of extracting a single transformer block, we collect patch features from a sparse set of layers,
$\mathcal{L}_{\mathrm{skip5}}=\{-1,-3,-6,-9,-12\}$.
For each selected layer $\ell\in\mathcal{L}_{\mathrm{skip5}}$, we obtain patch embeddings $Z^{(\ell)}\in\mathbb{R}^{P\times d}$, apply per-layer $\ell_2$ normalization to reduce scale mismatch, and concatenate them along the channel dimension:
\begin{equation}
Z_{\mathrm{cat}}=\mathrm{Concat}\big(\mathrm{Norm}(Z^{(\ell)})\big)_{\ell\in\mathcal{L}_{\mathrm{skip5}}}
\in \mathbb{R}^{P\times(|\mathcal{L}_{\mathrm{skip5}}|d)}.
\end{equation}
To control dimensionality, we apply a fixed random projection~\cite{achlioptas2003rp}:
\begin{equation}
Z = Z_{\mathrm{cat}}\mathbf{W},\quad
\mathbf{W}\in\mathbb{R}^{(|\mathcal{L}_{\mathrm{skip5}}|d)\times D},
\end{equation}
with $D=512$ by default and a projection matrix sampled once using a fixed seed.
We do not apply additional $\ell_2$ normalization after projection to preserve the distance scale used by kNN scoring.

\subsection{Category Memory Bank via Coreset Selection}
\label{sec:method:memory}
For each category $c$, we collect patch embeddings from normal training images into a set
$\mathcal{Z}_c=\{z_i\}_{i=1}^{T_c}$.
Storing all patches is infeasible, so we compress them into a memory bank using an approximate greedy farthest-point strategy (k-center style) as in PatchCore~\cite{sener2018coresets,gonzalez1985kcenter,roth2022patchcore}.
The coreset ratio $p$ is set by the experimental protocol (e.g., 1\% or 10\%).
For efficiency, distances for coreset selection can be computed in a low-dimensional proxy space, while the final stored embeddings remain $D$-dimensional.
The resulting scoring memory bank is stored as $\mathbf{M}_c\in\mathbb{R}^{N_c\times D}$.

\subsection{Routed Memory Inference for MUAD}
\label{sec:method:routing}
We consider multi-category UAD (MUAD), where a single deployed system must handle multiple categories $\mathcal{C}$ under a unified inference interface.
In this setting, exhaustive scoring against all category memory banks scales linearly with $|\mathcal{C}|$ and increases deployment cost.
To reduce this overhead, we use distance-based routing to bind each input to a single category and then query only the corresponding memory bank $\mathbf{M}_{\hat{c}}$.
At test time, the category identity is not assumed to be given, so the system must both select a relevant category memory and compute the anomaly score.

For each category $c$, we construct a routing bank $\mathbf{R}_c\in\mathbb{R}^{K_r\times D}$ consisting of $K_r$ prototypes (default $K_r{=}64$).
We $\ell_2$-normalize embeddings for routing; $\tilde{z}$ and $\tilde{r}$ denote normalized patch embeddings and prototypes, respectively.
In this normalized space, we compute the nearest-prototype distance for each patch and aggregate over patches to obtain a routing score:
\begin{equation}
d_{p,c}=\min_{r\in\mathbf{R}_c}\lVert \tilde{z}_p-\tilde{r}\rVert_2^2,
\qquad
\mathrm{route}(c)=\frac{1}{P}\sum_{p=1}^{P} d_{p,c}.
\end{equation}
\begin{equation}
\hat{c}=\arg\min_{c\in\mathcal{C}} \mathrm{route}(c).
\end{equation}
GCR~\cite{chae2026gcr} formulates routing via such a distance-based score.
We follow this formulation to build category-wise prototype banks and use routing to select a single scoring memory bank, thereby eliminating exhaustive scoring across categories.

\subsection{kNN-based Patch Anomaly Scoring and Base Decision}
\label{sec:method:knn}
Given the routed category $\hat{c}$, we perform patch-wise kNN search against $\mathbf{M}_{\hat{c}}$.
The patch anomaly score is defined as the mean squared $\ell_2$ distance to the $k$ nearest neighbors:
\begin{equation}
s_p=\frac{1}{k}\sum_{i=1}^{k}\lVert z_p-m_{p,i}\rVert_2^2.
\end{equation}
Patch scores $\{s_p\}_{p=1}^{P}$ are reshaped into a grid to form an anomaly map $S$, which is then upsampled and optionally smoothed with a Gaussian filter to match image resolution.
The base image-level score is defined by a pooling operator $\mathcal{P}$ applied to the anomaly map:
\begin{equation}
S_{\mathrm{base}}(x)=\mathcal{P}(S(x)).
\end{equation}
A common default is max pooling~\cite{roth2022patchcore,defard2021padim,cohen2020spade}.
StructCore complements this reduction by incorporating distributional and spatial structure that pooling discards.

\subsection{StructCore: Structure-aware Image-level Scoring}
\label{sec:method:structcore}
Max pooling retains only a single extreme response from the anomaly map and ignores how evidence is distributed and organized spatially.
StructCore addresses this by summarizing the distributional and spatial structure of an anomaly map using a minimal low-dimensional descriptor, and measuring its deviation under train-good statistics to refine the image-level decision.

Let $S\in\mathbb{R}^{H\times W}$ denote an anomaly map and $s\in\mathbb{R}^{HW}$ its flattened vector.
We use a \textbf{3D structural descriptor} $\phi(S)$ that captures complementary cues beyond the maximum response:
(i) global dispersion, (ii) tail concentration, and (iii) spatial roughness:
\begin{equation}
\phi(S)=\big[\ \sigma_S,\ \mathrm{topk\_mean}_r,\ \mathrm{TV}(S)\ \big]\in\mathbb{R}^{3},
\end{equation}
where $\sigma_S$ is the standard deviation of $s$.
To represent tail mass robustly, we use the mean of the top-$k$ scores with a fixed ratio $r$:
\begin{equation}
k=\max(1,\lfloor HW\cdot r\rfloor),\qquad
\mathrm{topk\_mean}_r=\frac{1}{k}\sum_{i\in \mathrm{Top}\text{-}k(s)} s_i,
\end{equation}
and quantify spatial roughness via total variation:
\begin{equation}
\mathrm{TV}(S)=\frac{1}{HW}\sum_{i,j}\big(|S_{i+1,j}-S_{i,j}|+|S_{i,j+1}-S_{i,j}|\big).
\end{equation}
We set $r=0.01$ (i.e., top 1\%) by default.

\paragraph{Train-good statistical calibration.}
We fit normal statistics of $\phi(S)$ using train-good samples only.
Let $\mu\in\mathbb{R}^{3}$ and $\sigma\in\mathbb{R}^{3}$ denote the per-dimension mean and standard deviation of $\phi(S)$ over train-good anomaly maps.
At test time, we compute a diagonal Mahalanobis (i.e., standardized Euclidean) distance:
\begin{equation}
D_{\mathrm{struct}}(S)=\left\lVert \frac{\phi(S)-\mu}{\sigma+\varepsilon}\right\rVert_2 .
\end{equation}
This structural score depends only on the anomaly map $S$ and does not modify patch scores or localization.

\paragraph{Automatic scale matching (auto-$\lambda$).}
Because $S_{\mathrm{base}}$ and $D_{\mathrm{struct}}$ can have different scales, we compute a \emph{category-wise} automatic weight using train-good statistics:
\begin{equation}
\lambda_{\mathrm{auto}}=\frac{\mathrm{Std}(S_{\mathrm{base}})}{\mathrm{Std}(D_{\mathrm{struct}})+\varepsilon},
\end{equation}
where the standard deviations are computed over train-good samples.
The final hybrid image-level score is
\begin{equation}
S_{\mathrm{hyb}}(x)=S_{\mathrm{base}}(x)+\lambda_{\mathrm{auto}}\cdot D_{\mathrm{struct}}(S(x)).
\end{equation}
For analysis, we also report a \textit{struct-only} score $S_{\mathrm{struct}}(x)=D_{\mathrm{struct}}(S(x))$.
StructCore modifies only the image-level score and preserves pixel-level localization behavior.

\section{Experiments}

\subsection{Experimental Settings}
\label{sec:exp:setup}

\paragraph{Benchmarks and Protocols.}
We evaluate StructCore on two standard industrial UAD benchmarks, \textbf{MVTec AD} (15 categories) and \textbf{VisA} (12 categories).
Following the one-class protocol, all components are fitted using only train-good samples, and evaluation is performed on test sets containing both normal and anomalous images.

\paragraph{Evaluation Metrics.}
We report both detection and localization metrics.
For \textbf{image-level} detection, we report I-AUROC and I-AP, and additionally report I-F1-max (the best F1 over thresholds) as a reference.
For \textbf{pixel-level} localization, we report P-AUROC, P-AP, and P-F1-max when pixel annotations are available, and additionally report AUPRO on MVTec AD.

\paragraph{Implementation and Backbone Details.}
\textbf{Feature extraction:} We use a frozen DINOv2 ViT-B/14 backbone (patch size 14)~\cite{oquab2023dinov2}.
\textbf{Preprocessing:} We follow the preprocessing protocol of Dinomaly~\cite{guo2025dinomaly}: images are resized to $448\times448$ and center-cropped to $392\times392$, producing a $28\times28$ token grid.
\textbf{Base detector:} We use a PatchCore-style memory-based detector with skip-layer fusion, coreset selection, and kNN retrieval.
\textbf{StructCore:} StructCore operates on anomaly maps without gradient-based training and refines only the image-level score (the anomaly map is unchanged).
We use the 3D descriptor $\phi(S)=\{\sigma_S,\mathrm{topk\_mean}_r,\mathrm{TV}\}$ (Sec.~\ref{sec:method:structcore}) and compute $D_{\mathrm{struct}}$ via a diagonal Mahalanobis distance.
To avoid tuning, we use an automatic scale calibration $\lambda_{\mathrm{auto}}$ estimated from train-good statistics (Sec.~\ref{sec:method:structcore}).

All experiments are conducted on a single NVIDIA RTX 3090 GPU (24GB).
We use FAISS-GPU for efficient nearest-neighbor search~\cite{johnson2017faiss} when available.
All stochastic components (random projection and coreset selection) use a fixed global seed (\texttt{SEED=42}) for reproducibility.
We use a \textbf{1\% coreset} as the default (memory-efficient) setting.
Table~\ref{tab:default_config} summarizes the paper-default configuration.

\begin{table}[t]
\centering
\small
\caption{Paper-default configuration. We use a 1\% coreset by default and report 10\% as a robustness check.}
\label{tab:default_config}
\setlength{\tabcolsep}{8pt}
\renewcommand{\arraystretch}{1.08}
\begin{tabular}{l l}
\hline
\textbf{Component} & \textbf{Setting} \\
\hline
Backbone & DINOv2 ViT-B/14 (frozen) \\
Input & resize $448\times448$, center-crop $392\times392$ \\
Layers & skip5: $[-1,-3,-6,-9,-12]$ (concat fusion) \\
Projection & fixed random projection to $D{=}512$ (RP-512) \\
Memory bank & coreset ratio = 1\% (default), 10\% (robustness) \\
Patch scoring & kNN $k{=}5$, mean over neighbors (squared $\ell_2$) \\
Map smoothing & Gaussian blur $\sigma{=}4$ \\
Pooling $\mathcal{P}$ & max pooling for the base score unless stated otherwise \\
StructCore & 3D $\phi(S)=\{\sigma_S,\mathrm{topk\_mean}_r,\mathrm{TV}\}$; diagonal Mahalanobis; auto-$\lambda$ (train-good statistics) \\
Routing & bank size $K_r{=}64$ per category; aggregation = mean over patch-to-prototype distances \\
\hline
\end{tabular}
\end{table}

\subsection{Comparison}
\label{sec:exp:comparison}

\paragraph{Baselines and comparison protocols.}
We compare \textbf{StructCore} under two complementary settings.
First, in the class-separated setting on MVTec AD (category known at inference), we compare against representative memory-bank / retrieval-based methods including CFA~\cite{cfa}, PaDiM~\cite{defard2021padim}, and PatchCore~\cite{roth2022patchcore}.
Second, to reflect scalable deployment, we also report results in a unified multi-category setting on MVTec AD and VisA and include recent unified approaches (RD4AD, DeSTSeg, UniAD, ReContrast, DiAD, ViTAD, MambaAD, Dinomaly) as a reference range.

Table~\ref{tab:mvtec_visa_mean_summary} summarizes the class-separated MVTec AD results by backbone group, reporting mean performance under the \textbf{1\% coreset} setting.
Within the WideResNet-50 group, PatchCore achieves 98.9/98.1 (I-AUROC/P-AUROC), and applying StructCore-style image scoring improves image-level AUROC to 99.1--99.2 while pixel-level AUROC remains unchanged.
Within the DINOv2 group, our base pipeline (max pooling on the same anomaly maps) attains 98.7/98.1, and enabling StructCore improves image-level AUROC to 99.6 with identical pixel-level localization metrics, indicating a pure decision-layer gain.

A per-category breakdown is provided in the Appendix for both \textbf{1\%} and \textbf{10\%} coresets
(Tables~\ref{tab:mvtec_1pct_detailed},~\ref{tab:mvtec_10pct_detailed},~\ref{tab:visa_1pct_detailed},~\ref{tab:visa_10pct_detailed}).
Our analysis shows that on MVTec AD (1\%), StructCore improves the mean image-level AUROC from 98.7 to 99.6.
Notable gains are observed on challenging categories, including Pill (+5.4), Screw (+2.7), Capsule (+2.2), and Cable (+1.8).
The worst-category image-level AUROC increases from 93.7 (Base) to 97.8 (Hybrid).
On VisA (1\%), the mean image-level AUROC increases from 97.6 to 98.4, with sizable improvements on Cashew (+4.0) and PCB1 (+2.0).
A small number of categories show minor decreases (typically around 0.1--0.3 AUROC), and the magnitude of these drops remains limited.

Finally, Table~\ref{tab:mvtec_visa_perf_merged_nomultirow} places our approach in the context of recent unified multi-category methods, reporting mean performance under the \textbf{1\% coreset} setting.
On MVTec AD, our method is competitive with strong trained baselines and achieves high image-level performance while maintaining strong localization.
On VisA, we obtain 98.4/98.5/95.7 at image level and 98.6/50.1/53.4 with AUPRO 95.6 at pixel level (1\% coreset), providing strong localization and competitive detection in a scalable setting.
Overall, these results support StructCore as a drop-in image-level decision layer that improves reliability without altering pixel-level localization behavior.

\begin{figure}[t]
  \centering
  \begin{minipage}{0.49\linewidth}
    \centering
    \includegraphics[width=\linewidth]{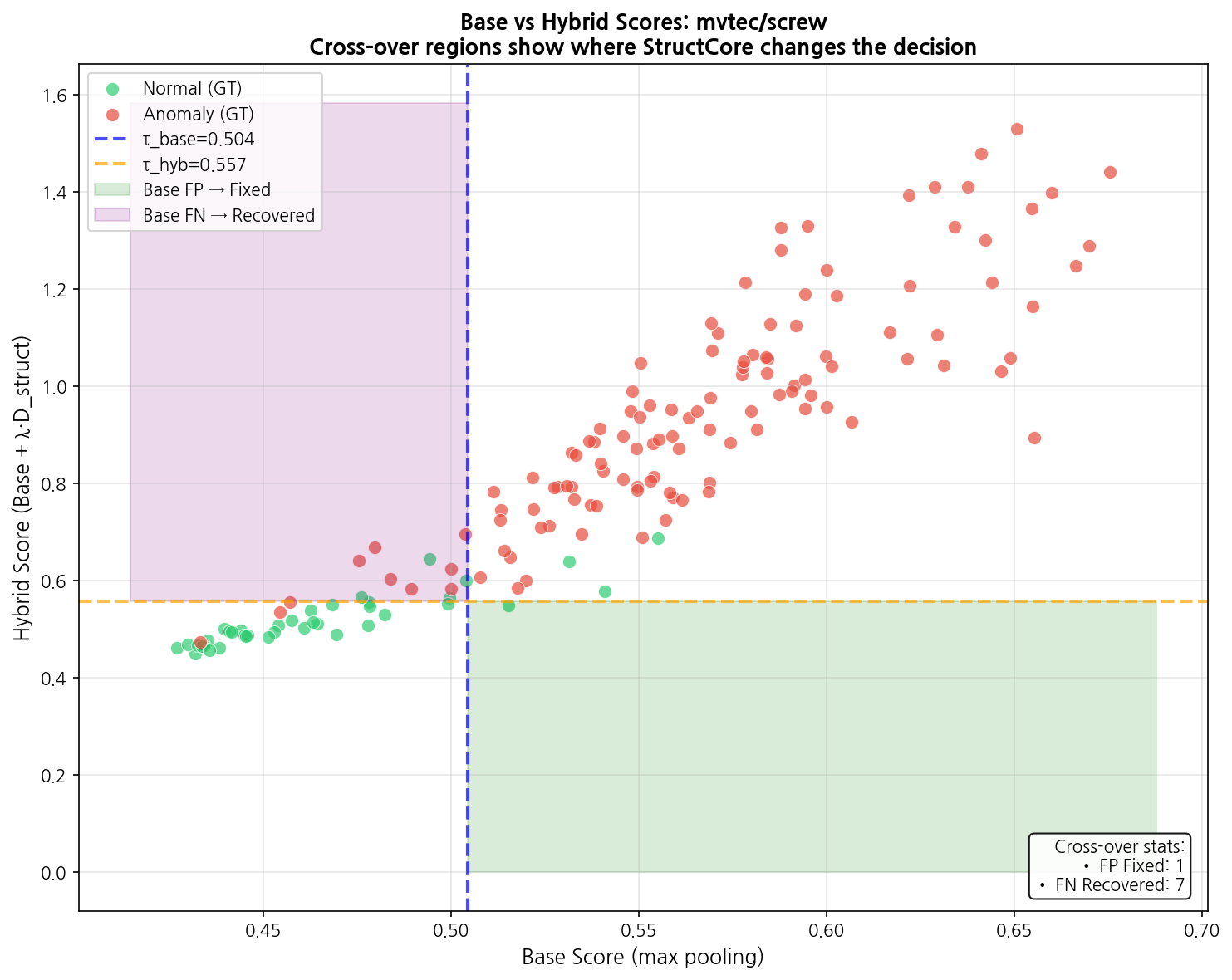}
    {\small (a) MVTec AD / \textit{screw}}
  \end{minipage}
  \hfill
  \begin{minipage}{0.49\linewidth}
    \centering
    \includegraphics[width=\linewidth]{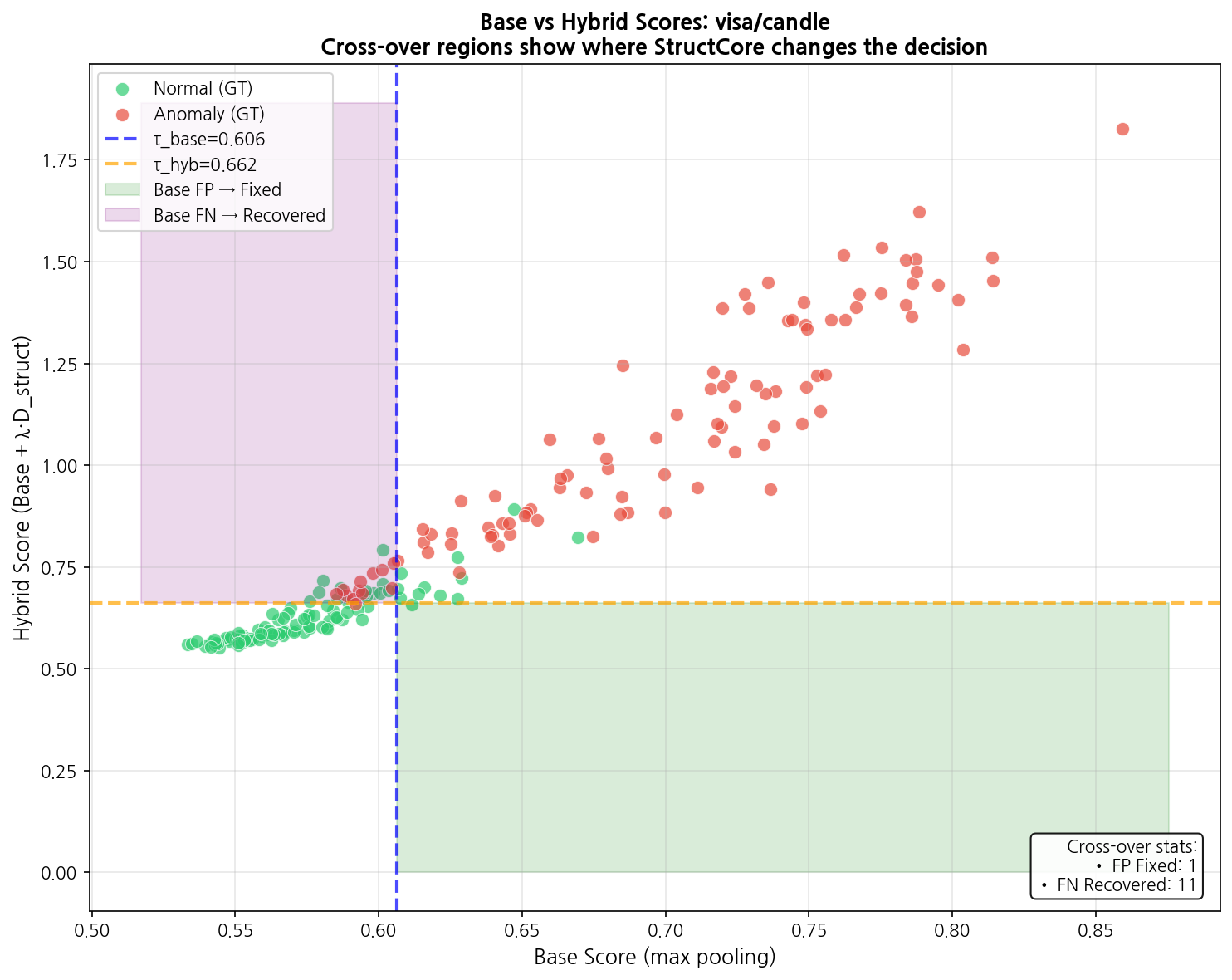}
    {\small (b) VisA / \textit{candle}}
  \end{minipage}

  \caption{Base vs. hybrid image scores illustrating where StructCore changes the image-level decision.
  The x-axis is the base score $S_{\mathrm{base}}=\max(S)$ and the y-axis is the final score $S_{\mathrm{hyb}}=S_{\mathrm{base}}+\lambda_{\mathrm{auto}}\,D_{\mathrm{struct}}$.
  Points are colored by ground-truth label.
  Thresholds $\tau_{\mathrm{base}}$ and $\tau_{\mathrm{hyb}}$ are set to the 99.5\% quantile of train-good scores.
  Shaded cross-over regions mark samples whose decisions differ between $S_{\mathrm{base}}$ and $S_{\mathrm{hyb}}$ (counts are shown in each panel).}
  \label{fig:scatter_crossover}
\end{figure}

\FloatBarrier

\begin{table*}[t]
\centering
\normalsize
\caption{Average performance (Mean) on MVTec-AD. Each entry reports \textbf{I-AUROC / P-AUROC}. 
\textbf{Results are only comparable within each backbone group.}
CFA/PaDiM/PatchCore follow their standard WideResNet-50 settings, while our variants use a frozen DINOv2 ViT-B/14 backbone with skip-layer fusion.
Within the DINOv2 group, \textit{Base} denotes max pooling on the same anomaly maps, and \textit{Struct/Hybrid} modify only the image-level scoring (pixel-level metrics remain unchanged).
For PatchCore-based runs in this table, the coreset ratio is set to 1\%.
Best results are highlighted in bold within each backbone group.}
\label{tab:mvtec_visa_mean_summary}

\setlength{\tabcolsep}{6pt}
\renewcommand{\arraystretch}{1.15}

\begin{tabular}{l l l cc}
\toprule
Dataset & Method & Backbone & I-AUROC & P-AUROC \\
\midrule

\multirow{8}{*}{MVTec-AD~\cite{bergmann2019mvtec}}
& CFA~\cite{cfa} & WideResNet-50 & 99.2 & 98.2 \\
& PaDiM~\cite{defard2021padim} & WideResNet-50 & 95.5 & 97.3 \\
& PatchCore (max pooling)~\cite{roth2022patchcore} & WideResNet-50 & 98.9 & 98.1 \\
& PatchCore + StructCore (struct only) & WideResNet-50 & 99.1 & 98.1 \\
& PatchCore + StructCore (hybrid) & WideResNet-50 & \textbf{99.2} & 98.1 \\
\cmidrule(lr){2-5}
& Ours (DINOv2, max pooling) & DINOv2 ViT-B/14 (skip-layer) & 98.7 & 98.2 \\
& Ours + StructCore (struct only) & DINOv2 ViT-B/14 (skip-layer) & \textbf{99.6} & 98.1 \\
& Ours + StructCore (hybrid, auto-$\lambda$) & DINOv2 ViT-B/14 (skip-layer) & \textbf{99.6} & 98.1 \\
\bottomrule
\end{tabular}
\end{table*}


\begin{table*}[t]
\centering
\small
\caption{Performance (\%) on MVTec-AD and VisA. Our StructCore results are reported with a \textbf{1\%} coreset. Best results are in bold.}

\label{tab:mvtec_visa_perf_merged_nomultirow}
\setlength{\tabcolsep}{3.5pt}
\renewcommand{\arraystretch}{1.1}
\resizebox{\textwidth}{!}{%
\begin{tabular}{l l ccc ccc c}
\toprule
Dataset & Method
& \multicolumn{3}{c}{Image-level}
& \multicolumn{3}{c}{Pixel-level}
& AUPRO \\
\cmidrule(lr){3-5}\cmidrule(lr){6-8}
& & I-AUROC & I-AP & I-F1-max & P-AUROC & P-AP & P-F1-max & \\
\midrule

MVTec-AD~\cite{bergmann2019mvtec}
& RD4AD~\cite{deng2022anomaly}                 & 94.6 & 96.5 & 95.2 & 96.1 & 48.6 & 53.8 & 91.1 \\
& SimpleNet~\cite{simplenet}                 & 95.3 & 98.4 & 95.8 & 96.9 & 45.9 & 49.7 & 86.5 \\
& DeSTSeg~\cite{destseg}                     & 89.2 & 95.5 & 91.6 & 93.1 & 54.3 & 50.9 & 64.8 \\
& UniAD~\cite{uniad}                         & 96.5 & 98.8 & 96.2 & 96.8 & 43.4 & 49.5 & 90.7 \\
& ReContrast~\cite{recontrast}               & 98.3 & 99.4 & 97.6 & 97.1 & 60.2 & 61.5 & 93.2 \\
& DiAD~\cite{diad}                           & 97.2 & 99.0 & 96.5 & 96.8 & 52.6 & 55.5 & 90.7 \\
& ViTAD~\cite{vitad}                         & 98.3 & 99.4 & 97.3 & 97.7 & 55.3 & 58.7 & 91.4 \\
& MambaAD~\cite{mambaad}                     & 98.6 & 99.6 & 97.8 & 97.7 & 56.3 & 59.2 & 93.1 \\
& Dinomaly~\cite{guo2025dinomaly}            & \textbf{99.6} & 99.8 & 99.0 & \textbf{98.4} & 69.3 & \textbf{69.2} & 94.8 \\
& StructCore (Ours, Base, 1\%)               & 98.7 & 99.6 & 98.1 & 98.1 & 68.7 & 67.9 & 94.2\\
& StructCore (Ours, Hybrid, 1\%)             & \textbf{99.6} & \textbf{99.9} & \textbf{99.4} & 98.1 & 68.7 & 67.9 & 94.2\\
\midrule

VisA~\cite{zou2022spot}
& RD4AD~\cite{deng2022anomaly}                 & 92.4 & 92.4 & 89.6 & 98.1 & 38.0 & 42.6 & 91.8 \\
& DeSTSeg~\cite{destseg}                     & 88.9 & 89.0 & 85.2 & 96.1 & 39.6 & 43.4 & 67.4 \\
& UniAD~\cite{uniad}                         & 88.8 & 90.8 & 85.8 & 98.3 & 33.7 & 39.0 & 85.5 \\
& ReContrast~\cite{recontrast}               & 95.5 & 96.4 & 92.0 & \textbf{98.5} & 47.9 & 50.6 & 91.9 \\
& DiAD~\cite{diad}                           & 86.8 & 88.3 & 85.1 & 96.0 & 26.1 & 33.0 & 75.2 \\
& ViTAD~\cite{vitad}                         & 90.5 & 91.7 & 86.3 & 98.2 & 36.6 & 41.1 & 85.1 \\
& MambaAD~\cite{mambaad}                     & 94.3 & 94.5 & 89.4 & \textbf{98.5} & 39.4 & 44.0 & 91.0 \\
& Dinomaly~\cite{guo2025dinomaly}            & \textbf{98.7} & \textbf{98.9} & \textbf{96.2} & \textbf{98.7} & \textbf{53.2} & 55.7 & 94.5 \\
& StructCore (Ours, Base, 1\%)               & 97.6 & 98.1 & 94.3 & 98.6 & 50.1 & 53.4 & \textbf{95.6} \\
& StructCore (Ours, Hybrid, 1\%)             & 98.4 & 98.5 & 95.7 & 98.6 & 50.1 & 53.4 & \textbf{95.6} \\

\bottomrule
\end{tabular}%
}
\end{table*}

\begin{table}[t]
\centering
\small
\caption{Scalability with memory-bank size on MVTec AD (RTX 3090, batch=16). 
``Extract'' includes DINOv2 feature extraction and random projection; ``kNN'' is FAISS search; ``Post'' includes Gaussian blur and StructCore scoring.
Coreset ratio controls the memory-bank size.}
\label{tab:latency_coreset}
\setlength{\tabcolsep}{7pt}
\renewcommand{\arraystretch}{1.08}
\begin{tabular}{c c c c c c c}
\toprule
Coreset & Extract (ms) & kNN (ms) & Post (ms) & Total (ms) & FPS & GPU peak (MB) \\
\midrule
1\%  & 9.60 & 0.20 & 0.32 & 10.12 & 98.85 & 2333 \\
5\%  & 9.68 & 0.68 & 0.31 & 10.67 & 93.71 & 2459 \\
10\% & 9.73 & 1.29 & 0.27 & 11.30 & 88.53 & 2619 \\
\bottomrule
\end{tabular}
\end{table}

\subsection{Ablation Study}
\label{sec:exp:ablation}

\paragraph{Sensitivity to the hybrid weight $\lambda$.}
StructCore uses the hybrid weight $\lambda$ to control the contribution of the structural calibration term.
We perform a sensitivity analysis over a fixed $\lambda \in \{0.05, 0.10, 0.15, 0.20\}$ across all categories.
While the optimal $\lambda$ may vary by category, the overall mean image-level AUROC remains essentially unchanged (around 99.6\%) across these values, indicating that StructCore is not overly sensitive to $\lambda$ in this range.
Unless stated otherwise, we use the automatic weight $\lambda_{\mathrm{auto}}$ in all experiments to avoid any tuning.

\paragraph{structural descriptor $\phi(S)$.}
StructCore hinges on a compact structural descriptor $\phi(S)$ that summarizes an anomaly score map beyond its maximum response.
Because $\phi(S)$ is intentionally low-dimensional and training-free, it is important to verify that
(i) each component captures complementary information that max pooling discards, and
(ii) the full 3D descriptor provides consistent benefits over simpler subsets under the same base detector.

Table~\ref{tab:phi_feature_ablation_mean} performs a controlled ablation by fixing the base detector and varying only $\phi(S)$.
Each component alone yields a clear improvement over the base image score (e.g., $\{\sigma_S\}$: +0.89, $\{\mathrm{topk\_mean}_r\}$: +0.88), indicating that both global dispersion and tail evidence provide useful calibration signals.
Combining components is consistently beneficial, and the full 3D descriptor $\{\sigma_S,\mathrm{topk\_mean}_r,\mathrm{TV}\}$ achieves the best performance (+0.99), supporting our design choice of a minimal yet effective structural basis.

\begin{table}[t]
\centering
\normalsize
\caption{Ablation of $\phi(S)$ components on MVTec AD (mean I-AUROC, \%).
Base uses max pooling; Hybrid uses $S_{\mathrm{hyb}} = S_{\mathrm{base}} + \lambda_{\mathrm{auto}} D_{\mathrm{struct}}$.
$\Delta$ denotes the improvement over Base.}
\label{tab:phi_feature_ablation_mean}
\setlength{\tabcolsep}{6pt}
\renewcommand{\arraystretch}{1.05}
\begin{tabular}{lccc}
\toprule
$\phi(S)$ & Base & Hybrid & $\Delta$ \\
\midrule
-- (Base only) & 98.68 & --    & -- \\
$\{\sigma_S\}$ & 98.68 & 99.57 & +0.89 \\
$\{\mathrm{topk\_mean}_r\}$ & 98.68 & 99.57 & +0.88 \\
$\{\mathrm{TV}\}$ & 98.68 & 99.07 & +0.38 \\
$\{\sigma_S,\mathrm{topk\_mean}_r\}$ & 98.68 & 99.66 & +0.97 \\
$\{\sigma_S,\mathrm{TV}\}$ & 98.68 & 99.55 & +0.87 \\
$\{\mathrm{topk\_mean}_r,\mathrm{TV}\}$ & 98.68 & 99.59 & +0.90 \\
$\{\sigma_S,\mathrm{topk\_mean}_r,\mathrm{TV}\}$ & 98.68 & \textbf{99.67} & \textbf{+0.99} \\
\bottomrule
\end{tabular}
\end{table}

\paragraph{Choice of structural distance for $D_{\mathrm{struct}}$.}
StructCore computes a structural score by measuring how atypical $\phi(S)$ is relative to train-good statistics.
While we use the diagonal Mahalanobis distance by default, other distance functions are possible.
To assess sensitivity to this choice, we compare several distances (e.g., $\ell_1$, $\ell_2$, cosine, and Chebyshev) as well as their standardized variants that normalize each dimension by the train-good standard deviation.
Table~\ref{tab:struct_distance_ablation} reports mean image-level AUROC (mean $\pm$ std over categories).
Across both datasets, standardized distances perform consistently well, and diagonal Mahalanobis remains competitive, indicating that StructCore is not overly sensitive to the particular distance choice.
Unless otherwise stated, we use the diagonal Mahalanobis distance for simplicity and for consistency with prior calibration literature.

\begin{table}[t]
\centering
\small
\caption{Ablation of structural distance for computing $D_{\mathrm{struct}}$ (mean image-level AUROC, mean $\pm$ std over categories).
Standardized variants (suffix \texttt{\_std}) normalize each dimension of $\phi(S)$ by the train-good standard deviation.}
\label{tab:struct_distance_ablation}
\setlength{\tabcolsep}{7pt}
\renewcommand{\arraystretch}{1.05}
\begin{tabular}{lcc}
\toprule
Distance & MVTec AD (15) & VisA (12) \\
\midrule
$\ell_1$ (std)    & \textbf{0.9965} $\pm$ 0.0057 & 0.9743 $\pm$ 0.0347 \\
Diagonal Mahalanobis & 0.9962 $\pm$ 0.0059 & 0.9837 $\pm$ 0.0319 \\
Chebyshev (std)   & 0.9960 $\pm$ 0.0060 & \textbf{0.9842} $\pm$ 0.0293 \\
Manhattan         & 0.9934 $\pm$ 0.0108 & 0.9789 $\pm$ 0.0339 \\
Euclidean         & 0.9927 $\pm$ 0.0117 & 0.9767 $\pm$ 0.0349 \\
Chebyshev         & 0.9922 $\pm$ 0.0123 & 0.9761 $\pm$ 0.0351 \\
Cosine            & 0.9916 $\pm$ 0.0133 & 0.9752 $\pm$ 0.0356 \\
\bottomrule
\end{tabular}
\end{table}

\section{Conclusion}
We introduced \textbf{StructCore}, a training-free, structure-aware image-level scoring method for unsupervised anomaly detection that improves upon conventional max pooling by exploiting distributional and spatial signatures of anomaly score maps.
Across MVTec AD and VisA, StructCore consistently enhances image-level decision making while leaving pixel-level localization unchanged, making it a simple drop-in module for both class-separated and unified multi-category deployments.

\appendix
\section{Supplementary Material / Appendix}
\label{sec:supp}

\subsection{Detailed Analysis of Structural Components}
\label{sec:supp:ablation_analysis}

We provide a detailed per-category analysis of the ablation study discussed in the main text.
Table~\ref{tab:supp_phi_ablation_mvtec_transposed} reports the Image-level AUROC for each category in MVTec AD under different combinations of structural features within the structural descriptor $\phi(S)$.

The proposed StructCore significantly enhances detection in categories where the Base (max pooling) method typically struggles due to the lack of structural context.
For instance, in \textit{Cable} ($92.71\% \rightarrow 99.10\%$), \textit{Capsule} ($95.81\% \rightarrow 98.40\%$), and \textit{Screw} ($96.82\% \rightarrow 98.38\%$), the inclusion of structural features effectively captures anomalies that are missed by a single peak response.
This confirms that structural statistics are particularly vital for defects that are subtle or spatially distributed.
While individual components like global dispersion ($\{\sigma_S\}$) or tail-based statistics ($\{\mathrm{topk\_mean}_r\}$) alone provide substantial boosts over the baseline, the full combination ($\{\sigma_S, \mathrm{topk\_mean}_r, \mathrm{TV}\}$) achieves the highest mean AUROC of $99.67\%$.

This indicates that global dispersion, tail concentration, and spatial roughness capture complementary aspects of anomaly maps, leading to a more robust decision boundary. For categories where the Base model already achieves near-perfect performance (e.g., \textit{Bottle}, \textit{Hazelnut}, \textit{Leather}), the Hybrid scoring maintains $100.00\%$ accuracy.
This demonstrates that the structural calibration is safe and does not induce negative transfer or degradation in simpler cases where max pooling is already sufficient.

\begin{table*}[t]
\centering
\scriptsize
\caption{Per-category structural feature ablation on MVTec AD (image-level AUROC, \%). 
Columns show image-level scoring variants: Base is max pooling; others are Hybrid scores with the indicated $\phi(\cdot)$}
\label{tab:supp_phi_ablation_mvtec_transposed}
\setlength{\tabcolsep}{3.2pt}
\renewcommand{\arraystretch}{1.08}
\resizebox{\textwidth}{!}{%
\begin{tabular}{lcccccccc}
\toprule
Category
& Base
& $\{\sigma_S\}$
& $\{\mathrm{topk\_mean}_r\}$
& $\{\mathrm{TV}\}$
& $\{\sigma_S,\mathrm{topk\_mean}_r\}$
& $\{\sigma_S,\mathrm{TV}\}$
& $\{\mathrm{topk\_mean}_r,\mathrm{TV}\}$
& $\{\sigma_S,\mathrm{topk\_mean}_r,\mathrm{TV}\}$ \\

\midrule
Bottle      & 100.00 & 100.00 & 100.00 & 100.00 & 100.00 & 100.00 & 100.00 & 100.00 \\
Cable       & 92.71  & 99.04  & 98.41  & 94.77  & 99.23  & 98.97  & 98.35  & 99.10  \\
Capsule     & 95.81  & 98.52  & 98.21  & 98.88  & 98.28  & 98.60  & 98.32  & 98.40  \\
Hazelnut    & 100.00 & 100.00 & 100.00 & 100.00 & 100.00 & 100.00 & 100.00 & 100.00 \\
Metal Nut   & 100.00 & 100.00 & 100.00 & 99.76  & 100.00 & 99.90  & 100.00 & 100.00 \\
Pill        & 98.06  & 99.15  & 99.59  & 98.28  & 99.32  & 99.35  & 99.59  & 99.56  \\
Screw       & 96.82  & 97.21  & 98.20  & 96.72  & 98.30  & 96.80  & 98.28  & 98.38  \\
Toothbrush  & 100.00 & 100.00 & 100.00 & 100.00 & 100.00 & 100.00 & 100.00 & 100.00 \\
Transistor  & 98.58  & 99.96  & 99.96  & 99.71  & 100.00 & 99.96  & 100.00 & 100.00 \\
Zipper      & 99.89  & 99.89  & 99.95  & 99.89  & 99.95  & 99.87  & 99.95  & 99.89  \\
Carpet      & 100.00 & 100.00 & 100.00 & 99.96  & 100.00 & 100.00 & 100.00 & 100.00 \\
Grid        & 98.99  & 99.92  & 99.42  & 98.58  & 99.83  & 99.92  & 99.33  & 99.83  \\
Leather     & 100.00 & 100.00 & 100.00 & 100.00 & 100.00 & 100.00 & 100.00 & 100.00 \\
Tile        & 100.00 & 100.00 & 100.00 & 100.00 & 100.00 & 100.00 & 100.00 & 100.00 \\
Wood        & 99.39  & 99.91  & 99.82  & 99.47  & 99.91  & 99.91  & 100.00 & 99.91  \\
\midrule
Mean        & 98.68  & 99.57  & 99.57  & 99.07  & 99.66  & 99.55  & 99.59  & \textbf{99.67} \\
\bottomrule
\end{tabular}%
}
\end{table*}

\begin{table*}[t]
\centering
\caption{Detailed per-category performance on \textbf{MVTec AD (1\% coreset)}.
Best Image-level AUROC is highlighted in \textbf{bold}.}
\label{tab:mvtec_1pct_detailed}
\setlength{\tabcolsep}{2.5pt}
\renewcommand{\arraystretch}{1.1}
\scriptsize
\resizebox{\textwidth}{!}{%
\begin{tabular}{l | ccc | ccc | ccc | cccc}
\toprule
\multirow{2}{*}{Category} & \multicolumn{3}{c|}{Base (Max)} & \multicolumn{3}{c|}{Struct (Ours)} & \multicolumn{3}{c|}{Hybrid (Ours)} & \multicolumn{4}{c}{Pixel-level Metrics} \\
\cmidrule(lr){2-4} \cmidrule(lr){5-7} \cmidrule(lr){8-10} \cmidrule(lr){11-14}
& AUROC & AP & F1-max & AUROC & AP & F1-max & AUROC & AP & F1-max & AUROC & AP & F1-max & AUPRO \\
\midrule
Bottle      & \textbf{100.0} & 100.0 & 100.0 & \textbf{100.0} & 100.0 & 100.0 & \textbf{100.0} & 100.0 & 100.0 & 98.8 & 86.1 & 81.5 & 95.9 \\
Cable       & 97.7 & 98.7 & 93.9 & 99.4 & 99.7 & 97.3 & \textbf{99.5} & 99.7 & 97.2 & 94.5 & 52.6 & 62.5 & 88.0 \\
Capsule     & 95.6 & 99.0 & 96.8 & \textbf{97.8} & 99.4 & 98.6 & \textbf{97.8} & 99.5 & 98.2 & 98.3 & 59.7 & 56.8 & 94.5 \\
Carpet      & \textbf{100.0} & 100.0 & 100.0 & \textbf{100.0} & 100.0 & 100.0 & \textbf{100.0} & 100.0 & 100.0 & 99.3 & 73.4 & 70.4 & 97.8 \\
Grid        & 99.9 & 100.0 & 99.1 & 99.9 & 100.0 & 99.1 & \textbf{100.0} & 100.0 & 100.0 & 99.6 & 61.9 & 61.5 & 98.6 \\
Hazelnut    & \textbf{100.0} & 100.0 & 99.3 & \textbf{100.0} & 100.0 & 100.0 & \textbf{100.0} & 100.0 & 100.0 & 99.5 & 80.8 & 78.1 & 98.4 \\
Leather     & \textbf{100.0} & 100.0 & 100.0 & \textbf{100.0} & 100.0 & 100.0 & \textbf{100.0} & 100.0 & 100.0 & 99.3 & 50.1 & 51.0 & 97.8 \\
Metal Nut   & \textbf{100.0} & 100.0 & 100.0 & \textbf{100.0} & 100.0 & 100.0 & \textbf{100.0} & 100.0 & 100.0 & 97.8 & 82.9 & 85.0 & 92.8 \\
Pill        & 93.7 & 98.8 & 95.3 & \textbf{99.2} & 99.9 & 98.9 & \textbf{99.1} & 99.9 & 98.9 & 96.8 & 69.5 & 66.0 & 90.5 \\
Screw       & 95.6 & 98.4 & 94.9 & \textbf{98.6} & 99.5 & 96.7 & 98.3 & 99.5 & 97.1 & 99.5 & 64.7 & 61.7 & 98.4 \\
Tile        & \textbf{100.0} & 100.0 & 100.0 & \textbf{100.0} & 100.0 & 100.0 & \textbf{100.0} & 100.0 & 100.0 & 97.4 & 65.8 & 73.9 & 91.4 \\
Toothbrush  & 99.7 & 99.9 & 98.4 & \textbf{100.0} & 100.0 & 100.0 & \textbf{100.0} & 100.0 & 100.0 & 99.1 & 56.2 & 65.3 & 96.9 \\
Transistor  & 99.2 & 98.9 & 95.0 & \textbf{100.0} & 100.0 & 100.0 & \textbf{100.0} & 100.0 & 100.0 & 92.8 & 61.0 & 57.6 & 80.1 \\
Wood        & 99.6 & 99.9 & 99.2 & 99.9 & 100.0 & 99.2 & \textbf{100.0} & 100.0 & 100.0 & 97.7 & 76.3 & 71.3 & 93.2 \\
Zipper      & \textbf{99.9} & 100.0 & 99.6 & 99.8 & 99.9 & 99.6 & 99.9 & 100.0 & 99.6 & 98.1 & 66.6 & 64.3 & 94.1 \\
\midrule
\textbf{Mean} & 98.7 & 99.6 & 98.2 & \textbf{99.6} & 99.9 & 99.2 & \textbf{99.6} & 99.9 & 99.4 & 97.9 & 67.2 & 67.1 & 93.9 \\
\bottomrule
\end{tabular}%
}
\end{table*}

\begin{table*}[t]
\centering
\caption{Detailed per-category performance on \textbf{MVTec AD (10\% coreset)}.}
\label{tab:mvtec_10pct_detailed}
\setlength{\tabcolsep}{2.5pt}
\renewcommand{\arraystretch}{1.1}
\scriptsize
\resizebox{\textwidth}{!}{%
\begin{tabular}{l | ccc | ccc | ccc | cccc}
\toprule
\multirow{2}{*}{Category} & \multicolumn{3}{c|}{Base (Max)} & \multicolumn{3}{c|}{Struct (Ours)} & \multicolumn{3}{c|}{Hybrid (Ours)} & \multicolumn{4}{c}{Pixel-level Metrics} \\
\cmidrule(lr){2-4} \cmidrule(lr){5-7} \cmidrule(lr){8-10} \cmidrule(lr){11-14}
& AUROC & AP & F1-max & AUROC & AP & F1-max & AUROC & AP & F1-max & AUROC & AP & F1-max & AUPRO \\
\midrule
Bottle      & \textbf{100.0} & 100.0 & 100.0 & \textbf{100.0} & 100.0 & 100.0 & \textbf{100.0} & 100.0 & 100.0 & 99.0 & 87.1 & 81.0 & 96.5 \\
Cable       & 97.4 & 98.5 & 93.1 & 99.7 & 99.8 & 98.4 & \textbf{99.8} & 99.9 & 98.4 & 95.4 & 53.5 & 63.3 & 89.4 \\
Capsule     & 97.5 & 99.4 & 98.6 & 98.3 & 99.6 & 98.2 & \textbf{98.4} & 99.6 & 98.6 & 98.6 & 61.0 & 58.1 & 95.5 \\
Carpet      & \textbf{100.0} & 100.0 & 100.0 & \textbf{100.0} & 100.0 & 100.0 & \textbf{100.0} & 100.0 & 100.0 & 99.4 & 74.4 & 70.9 & 97.9 \\
Grid        & 99.2 & 99.7 & 99.1 & \textbf{99.9} & 100.0 & 99.1 & 99.8 & 99.9 & 99.1 & 99.6 & 63.7 & 63.5 & 98.7 \\
Hazelnut    & \textbf{100.0} & 100.0 & 100.0 & \textbf{100.0} & 100.0 & 100.0 & \textbf{100.0} & 100.0 & 100.0 & 99.5 & 81.7 & 78.9 & 98.5 \\
Leather     & \textbf{100.0} & 100.0 & 100.0 & \textbf{100.0} & 100.0 & 100.0 & \textbf{100.0} & 100.0 & 100.0 & 99.4 & 50.9 & 51.9 & 97.8 \\
Metal Nut   & \textbf{100.0} & 100.0 & 100.0 & \textbf{100.0} & 100.0 & 100.0 & \textbf{100.0} & 100.0 & 100.0 & 97.9 & 82.7 & 85.8 & 93.0 \\
Pill        & \textbf{99.4} & 99.9 & 98.6 & 98.8 & 99.8 & 97.9 & 99.2 & 99.9 & 98.2 & 96.8 & 69.7 & 67.2 & 90.2 \\
Screw       & 95.7 & 98.4 & 94.5 & \textbf{98.4} & 99.5 & 97.0 & \textbf{98.4} & 99.5 & 96.1 & 99.6 & 65.5 & 62.8 & 98.8 \\
Tile        & \textbf{100.0} & 100.0 & 100.0 & \textbf{100.0} & 100.0 & 100.0 & \textbf{100.0} & 100.0 & 100.0 & 97.5 & 65.7 & 74.4 & 91.6 \\
Toothbrush  & \textbf{100.0} & 100.0 & 100.0 & \textbf{100.0} & 100.0 & 100.0 & \textbf{100.0} & 100.0 & 100.0 & 99.2 & 58.3 & 66.7 & 97.3 \\
Transistor  & 99.1 & 98.7 & 94.1 & 99.8 & 99.7 & 97.6 & \textbf{100.0} & 100.0 & 100.0 & 95.3 & 65.5 & 60.7 & 85.5 \\
Wood        & 99.6 & 99.9 & 99.2 & \textbf{99.9} & 100.0 & 99.2 & 99.8 & 99.9 & 99.2 & 97.7 & 76.2 & 71.2 & 93.3 \\
Zipper      & 99.8 & 100.0 & 99.6 & 99.9 & 100.0 & 99.6 & \textbf{99.9} & 100.0 & 99.6 & 98.2 & 67.1 & 64.2 & 94.3 \\
\midrule
\textbf{Mean} & 99.2 & 99.6 & 98.2 & 99.6 & 99.9 & 99.2 & \textbf{99.7} & 99.9 & 99.3 & 98.2 & 68.2 & 68.0 & 94.6 \\
\bottomrule
\end{tabular}%
}
\end{table*}

\begin{table*}[t]
\centering
\caption{Detailed per-category performance on \textbf{VisA (1\% coreset)}.}
\label{tab:visa_1pct_detailed}
\setlength{\tabcolsep}{2.5pt}
\renewcommand{\arraystretch}{1.1}
\scriptsize
\resizebox{\textwidth}{!}{%
\begin{tabular}{l | ccc | ccc | ccc | cccc}
\toprule
\multirow{2}{*}{Category} & \multicolumn{3}{c|}{Base (Max)} & \multicolumn{3}{c|}{Struct (Ours)} & \multicolumn{3}{c|}{Hybrid (Ours)} & \multicolumn{4}{c}{Pixel-level Metrics} \\
\cmidrule(lr){2-4} \cmidrule(lr){5-7} \cmidrule(lr){8-10} \cmidrule(lr){11-14}
& AUROC & AP & F1-max & AUROC & AP & F1-max & AUROC & AP & F1-max & AUROC & AP & F1-max & AUPRO \\
\midrule
Candle      & 96.7 & 97.0 & 89.7 & \textbf{98.4} & 98.4 & 93.8 & 98.2 & 98.3 & 92.9 & 99.4 & 39.4 & 45.6 & 98.2 \\
Capsules    & \textbf{98.9} & 99.4 & 96.5 & 96.8 & 97.8 & 93.8 & 98.6 & 99.1 & 97.5 & 99.3 & 62.9 & 64.0 & 97.7 \\
Cashew      & 94.2 & 97.3 & 90.8 & 98.1 & 99.0 & 96.6 & \textbf{98.2} & 99.1 & 95.2 & 96.9 & 58.4 & 59.3 & 91.1 \\
Chewinggum  & 99.4 & 99.7 & 98.0 & 99.3 & 99.7 & 99.0 & \textbf{99.6} & 99.8 & 99.0 & 99.4 & 75.6 & 69.8 & 98.1 \\
Fryum       & 98.6 & 99.3 & 95.3 & 98.1 & 99.1 & 94.9 & \textbf{98.6} & 99.4 & 95.8 & 95.2 & 43.2 & 49.5 & 85.2 \\
Macaroni1   & \textbf{98.4} & 98.6 & 94.0 & 95.9 & 95.7 & 91.9 & \textbf{98.4} & 98.6 & 94.1 & 99.7 & 25.3 & 32.4 & 99.1 \\
Macaroni2   & \textbf{94.3} & 92.3 & 90.3 & 90.4 & 89.2 & 85.7 & 94.2 & 93.3 & 89.4 & 99.7 & 21.5 & 29.0 & 99.0 \\
PCB1        & 96.4 & 95.7 & 91.9 & \textbf{98.5} & 98.3 & 95.2 & 98.4 & 98.2 & 94.7 & 99.5 & 86.3 & 80.2 & 98.6 \\
PCB2        & 96.3 & 97.3 & 93.9 & \textbf{97.2} & 96.3 & 93.7 & \textbf{97.2} & 96.9 & 95.4 & 98.3 & 43.2 & 49.7 & 94.9 \\
PCB3        & 98.8 & 98.9 & 94.8 & 99.1 & 99.2 & 95.5 & \textbf{99.5} & 99.5 & 96.5 & 98.9 & 44.8 & 47.9 & 96.5 \\
PCB4        & 99.5 & 99.5 & 97.0 & 99.9 & 99.9 & 98.5 & \textbf{99.9} & 99.9 & 99.0 & 97.6 & 44.4 & 49.2 & 92.0 \\
Pipe Fryum  & \textbf{99.9} & 100.0 & 99.5 & 98.4 & 99.3 & 96.5 & 99.6 & 99.8 & 99.0 & 99.1 & 56.2 & 64.4 & 97.0 \\
\midrule
\textbf{Mean} & 97.6 & 97.9 & 94.3 & 97.5 & 97.7 & 94.6 & \textbf{98.4} & 98.5 & 95.7 & 98.6 & 50.1 & 53.4 & 95.6 \\
\bottomrule
\end{tabular}%
}
\end{table*}

\begin{table*}[t]
\centering
\caption{Detailed per-category performance on \textbf{VisA (10\% coreset)}.}
\label{tab:visa_10pct_detailed}
\setlength{\tabcolsep}{2.5pt}
\renewcommand{\arraystretch}{1.1}
\scriptsize
\resizebox{\textwidth}{!}{%
\begin{tabular}{l | ccc | ccc | ccc | cccc}
\toprule
\multirow{2}{*}{Category} & \multicolumn{3}{c|}{Base (Max)} & \multicolumn{3}{c|}{Struct (Ours)} & \multicolumn{3}{c|}{Hybrid (Ours)} & \multicolumn{4}{c}{Pixel-level Metrics} \\
\cmidrule(lr){2-4} \cmidrule(lr){5-7} \cmidrule(lr){8-10} \cmidrule(lr){11-14}
& AUROC & AP & F1-max & AUROC & AP & F1-max & AUROC & AP & F1-max & AUROC & AP & F1-max & AUPRO \\
\midrule
Candle      & 96.3 & 96.7 & 90.9 & \textbf{97.9} & 98.0 & 93.5 & \textbf{97.9} & 98.0 & 92.8 & 99.6 & 40.6 & 47.4 & 98.6 \\
Capsules    & \textbf{99.3} & 99.6 & 97.5 & 97.0 & 98.0 & 95.0 & 98.5 & 98.8 & 98.0 & 99.5 & 63.8 & 63.9 & 98.3 \\
Cashew      & 96.6 & 98.3 & 94.0 & 98.2 & 99.0 & 97.6 & \textbf{98.6} & 99.2 & 97.5 & 93.7 & 52.2 & 55.6 & 83.7 \\
Chewinggum  & \textbf{99.4} & 99.7 & 98.5 & 99.1 & 99.6 & 98.5 & \textbf{99.4} & 99.8 & 99.0 & 99.3 & 77.7 & 71.7 & 98.0 \\
Fryum       & \textbf{99.1} & 99.6 & 96.6 & 97.8 & 98.9 & 96.0 & \textbf{99.1} & 99.6 & 96.5 & 95.7 & 43.3 & 48.8 & 86.1 \\
Macaroni1   & \textbf{97.2} & 97.8 & 92.5 & 91.4 & 91.1 & 86.0 & 95.6 & 95.9 & 91.3 & 99.7 & 24.8 & 32.2 & 98.9 \\
Macaroni2   & \textbf{94.7} & 93.5 & 90.7 & 86.4 & 84.4 & 81.2 & 93.1 & 92.3 & 87.8 & 99.8 & 21.9 & 30.0 & 99.2 \\
PCB1        & 96.4 & 95.6 & 91.2 & 98.2 & 97.9 & 94.6 & \textbf{98.3} & 98.2 & 94.7 & 99.6 & 85.4 & 79.2 & 98.8 \\
PCB2        & 97.0 & 97.2 & 92.2 & 96.8 & 95.0 & 94.5 & \textbf{97.4} & 96.0 & 94.4 & 99.0 & 44.5 & 49.0 & 96.9 \\
PCB3        & 99.0 & 99.0 & 96.1 & 98.3 & 98.2 & 94.6 & \textbf{99.0} & 99.1 & 96.5 & 99.2 & 45.7 & 47.0 & 97.3 \\
PCB4        & 99.5 & 99.4 & 97.6 & 99.9 & 99.9 & 98.5 & \textbf{100.0} & 100.0 & 99.0 & 98.3 & 46.7 & 50.2 & 94.4 \\
Pipe Fryum  & \textbf{99.9} & 100.0 & 99.5 & 96.8 & 98.6 & 93.8 & 99.1 & 99.6 & 98.0 & 99.0 & 54.5 & 63.1 & 96.8 \\
\midrule
\textbf{Mean} & 97.9 & 98.1 & 94.8 & 96.5 & 96.5 & 93.6 & \textbf{98.0} & 98.0 & 95.5 & 98.5 & 50.1 & 53.2 & 95.6 \\
\bottomrule
\end{tabular}%
}
\end{table*}

\paragraph{Effect of transformer depth.}
Table~\ref{tab:single_layer_sweep} shows that single-layer ViT features are sensitive to depth.
Performance peaks at intermediate-to-late blocks (e.g., MVTec AD: 98.94 at layer $-4/-5$; VisA: 97.11 at layer $-4$), while the final block can substantially degrade performance on VisA (80.75 at layer $-1$).
This motivates our skip-layer fusion, which aggregates complementary cues across depths to reduce depth sensitivity under a fixed memory-efficient protocol.

\begin{table}[t]
\centering
\caption{Single-layer depth sweep on MVTec AD and VisA.
All settings are fixed: RP dim=256, $k$=5, with the indicated $\phi(\cdot)$
Images are resized to 448 and center-cropped to 392.
We report image-level AUROC (\%).}
\label{tab:single_layer_sweep}
\begin{tabular}{c|cc}
\toprule
Layer idx & MVTec (15) & VisA (12) \\
\midrule
-1  & 98.58 & 80.75 \\
-2  & 98.61 & 92.97 \\
-3  & 98.78 & 95.91 \\
-4  & \textbf{98.94} & \textbf{97.11} \\
-5  & 98.94 & 96.65 \\
-6  & 98.86 & 97.00 \\
-7  & 98.56 & 96.20 \\
-8  & 98.41 & 95.64 \\
-9  & 97.44 & 96.93 \\
-10 & 95.87 & 95.83 \\
-11 & 89.57 & 91.92 \\
-12 & 77.50 & 84.94 \\
\bottomrule
\end{tabular}
\end{table}

\begin{figure*}[p]
  \centering
  \includegraphics[
    width=\textwidth,
    height=0.86\textheight,
    keepaspectratio
  ]{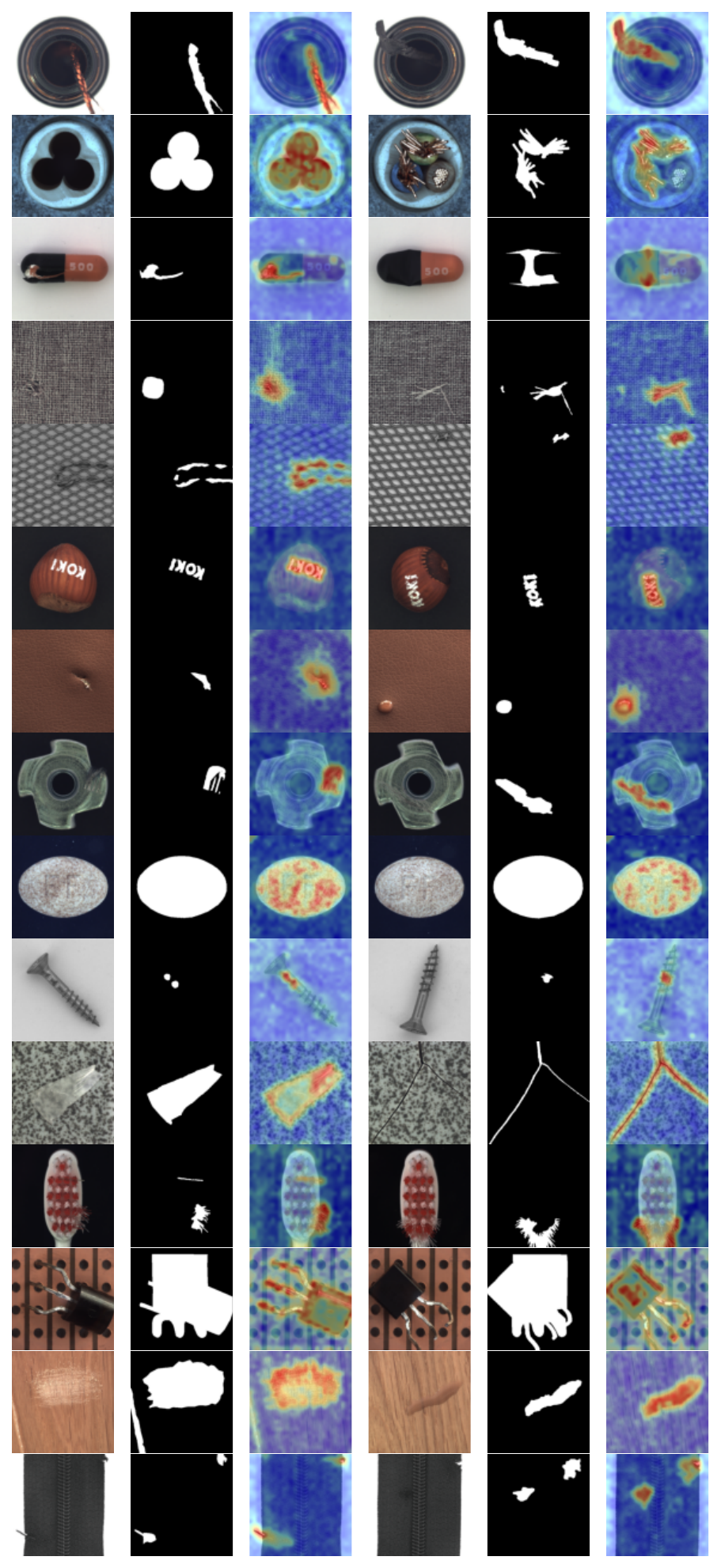}
  \caption{Qualitative results on MVTec AD (all categories).}
  \label{fig:qual_mvtec_allcats}
\end{figure*}

\begin{figure*}[p]
  \centering
  \includegraphics[
    width=\textwidth,
    height=0.86\textheight,
    keepaspectratio
  ]{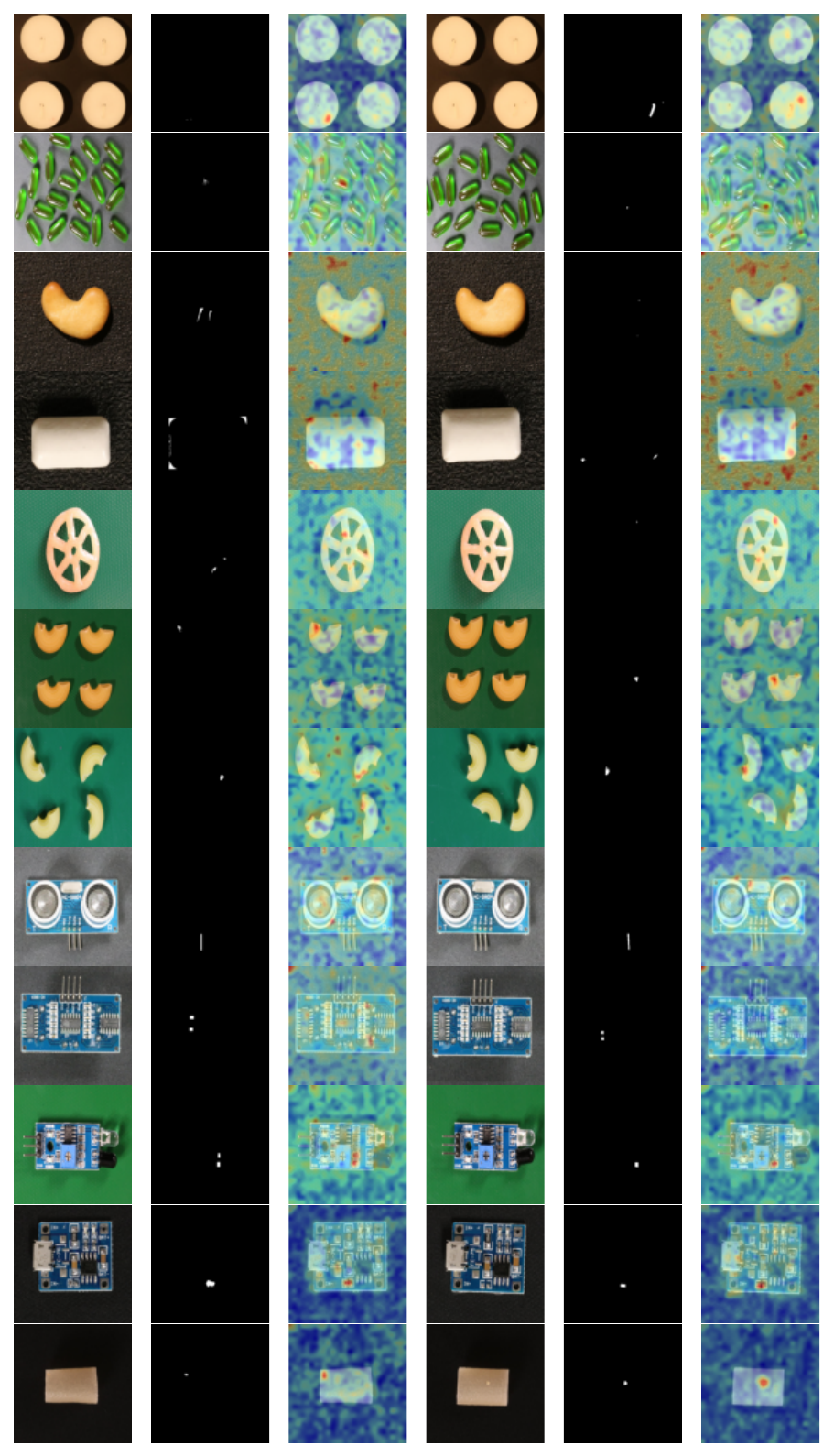}
  \caption{Qualitative results on VisA (all categories).}
  \label{fig:qual_visa_allcats}
\end{figure*}

\bibliographystyle{unsrt}  
\bibliography{references}

\end{document}